\documentclass[a4paper,conference]{IEEEtran}
\IEEEoverridecommandlockouts
\usepackage{cite}
\usepackage{amsmath,amssymb,amsfonts}
\usepackage{algorithmic}
\usepackage[dvipdfmx]{graphicx}
\usepackage{comment}
\usepackage{textcomp}
\usepackage{xcolor}
\usepackage{here}
\usepackage{mathtools}
\def\BibTeX{{\rm B\kern-.05em{\sc i\kern-.025em b}\kern-.08em
    T\kern-.1667em\lower.7ex\hbox{E}\kern-.125emX}}
\begin{document}

\title{Online Heterogeneous Mixture Learning \\for Big Data\\
{\footnotesize \textsuperscript{*}}
\thanks{}
}

\author{\IEEEauthorblockN{1\textsuperscript{st} Kazuki Seshimo}
\IEEEauthorblockA{\textit{} 
\textit{}
Kanazawa University, Kanazawa, Japan \\
kseshimo@csl.ec.t.kanazawa-u.ac.jp}
\and
\IEEEauthorblockN{2\textsuperscript{nd} Akira Ota}
\IEEEauthorblockA{\textit{} 
\textit{}
Kanazawa University, Kanazawa, Japan \\
aota@csl.ec.t.kanazawa-u.ac.jp}
\and
\IEEEauthorblockN{3\textsuperscript{rd} Daichi Nishio}
\IEEEauthorblockA{\textit{} 
\textit{}
Kanazawa University, Kanazawa, Japan \\
dnishio@csl.ec.t.kanazawa-u.ac.jp}
\and
\IEEEauthorblockN{4\textsuperscript{th} Satoshi Yamane}
\IEEEauthorblockA{\textit{} 
\textit{}
Kanazawa University, Kanazawa, Japan \\
syamane@is.t.kanazawa-u.ac.jp}

}

\maketitle

\begin{abstract}
We propose the online machine learning for big data analysis with heterogeneity.
We performed an experiment to compare the accuracy of each iteration between batch one and online one. 
It is possible to converge quickly with the same accuracy as the batch one.
\end{abstract}


\section{Introduction}
There is a kind of heterogeneous mixture learning for big data analysis with heterogeneity.
This is batch learning using a batch EM algorithm for model generation[1].
Therefore, we use the incremental EM algorithm [2,3,4] which is an online EM algorithm to propose online heterogeneous mixture learning.
Online heterogeneous mixture learning is possible to converge faster than the batch type with the same accuracy.
\section{online heterogeneous mixed learning}
We propose online learning of heterogeneous mixed learning using the online method of EM algorithm for mixture of Gaussian.
First of all, we introduce the incremental EM algorithm[2,3].\\
$E_{incremental}Step:$
We fix the parameters,
and calculate the responsibility
$\gamma$ and the amount of change in the responsibility
$s_k$.
we update the responsibility for one data $x_n$ with observation data
$x_N$.
  
\begin{equation}
  \gamma \left( z _ { n k } \right) ^ { ( t + 1 ) } = \frac { \pi _ { k } ^ { ( t ) }  N  \left( \boldsymbol { x } _ { n } | \boldsymbol { \mu } _ { k } ^ { ( t ) } , \boldsymbol { \Sigma } _ { k } ^ { ( t ) } \right) } { \sum _ { j = 1 } ^ { K } \pi _ { j } ^ { ( t ) }  N  \left( \boldsymbol { x } _ { n } | \boldsymbol { \mu } _ { j } ^ { ( t ) } , \boldsymbol { \Sigma } _ { j } ^ { ( t ) } \right) } , ( k = 1 , \ldots , K )
\end{equation}

We calculate the amount of change in the responsibility
$s_{nk}$.

  
\begin{equation}
  s _ { nk } ^ { ( t + 1 ) } = \gamma \left( z _ { n k } \right) ^ { ( t + 1 ) } - \gamma \left( z _ { n k } \right) ^ { ( t ) }
\end{equation}

\begin{equation}
  N _ { k } ^ { ( t + 1 ) } = \sum _ { n = 1 } ^ { N } \gamma \left( z _ { n k } \right) ^ { ( t  ) } + s _ { n k } ^ { ( t + 1 ) } = N _ { k } ^ { ( t  ) } + s _ { n k } ^ { ( t ) }
\end{equation}

\noindent
$M_{incremental}Step:$
We fix the esponsibility 
$\gamma(z_{nk})$ and the amount of change in the responsibility
$s_{nk}$,
and update each parameter.


\begin{equation}
  \pi _ { k } ^ { ( t + 1 ) } = \pi_{k}^{(t)} + \frac { s _ { nk } ^ { ( t + 1 ) } } { N }
\end{equation}
\begin{equation}
  \mu _ { k } ^ { ( t + 1 ) } = \mu _ { k } ^ { ( t ) } + \frac { s _ { nk } ^ { ( t + 1 ) } } { N _ { k } ^ { ( t + 1 ) } } \left( \boldsymbol { x } _ { n } - \boldsymbol { \mu } _ { k } ^ { ( t ) } \right)
\end{equation}
\begin{align}
  \mathbf { \Sigma } _ { k } ^ { ( t + 1 ) } &=\left( 1 - \frac { s _ { nk } ^ { ( t + 1 ) } } { N _ { k } ^ { ( t + 1 ) } } \right) \notag \\
  &\left\{ \mathbf { \Sigma } _ { k } ^ { ( t ) } + \frac { s _ { nk } ^ { ( t + 1 ) } } { N _ { k } ^ { ( t + 1 ) } } \left( \boldsymbol { x } _ { n } - \boldsymbol { \mu } _ { k } ^ { ( t ) } \right) \left( \boldsymbol { x } _ { n } - \boldsymbol { \mu } _ { k } ^ { ( t ) } \right) ^ { T } \right\}
\end{align}

The crucial points of heterogeneous mixed learning are a factorized information criterion (FIC) and factorized asymptotic Bayesian inference (FAB)[1].
We have to make these available online.
First, we improve FIC, which is metric of the model.
Second, we improve FAB in response to change of FIC.

The $FIC_{online}$ which supports online learning is shown below.

\begin{align}
  F I C_{\text {online}}\left(x_{n}, M\right)&=\operatorname{FIC}\left(\boldsymbol{x}^{N-1}, M\right)+F I C_{+}\left(x_{n}, M\right)\notag\\
  &=\max _{q}\left\{J_{\text {online}}\left(q, \overline{\theta}, x_{n}\right)\right\}
\end{align}
\begin{align}
  J_{\text { online }}\left(q, \overline{\theta}, x_{n}\right)&= J_{\text { online }}\left(q, \overline{\theta}, x_{n-1}\right)\notag \\
  &+q\left(z_{n c}\right)\left[\log p\left(x_{n}, z_{n c} | \overline{\theta}\right)\right.\notag\\
  & \left.-\frac{1}{2} \log N-\sum_{c=1}^{C} \frac{D_{c}}{2}\left\{\log z_{n c}-\log q\left(z_{n c}\right)\right\}\right] 
\end{align} 
It is not possible to evaluate $FIC_{online}$ directly because the parameters can not be determined analytically.
In order to evaluate FIC, FAB maximizes an asymptotically-consistent lower bound of FIC.
For updates incrementally, we improve FAB using the variation of the variational probability of the latent variable.

We calculate sequentially by repeating the following two steps $t$ times.\\\\
$V_{online}Step:$
We optimize the distribution
$q(z_{nc})$ of latent variables
${z}^N$, and calculate the distribution of latent variables and their variation
$s_{nc}$ for additional data
$x_n$.\\\\
$M_{online}Step:$
We optimize components of mixture of Gaussian and parameters
$\theta$.

\section{results of experiment}
We compare the results of conventional batch heterogeneous mixture learning [1] and online heterogeneous mixture learning which is proposed in this paper in the same environment and conditions.

In this experiment, the data used for learning is normal random number generated from the mixture of Gaussian.
The mixture of Gaussian needs three parameters which are means, convariances and mixing coefficient.
We specified these three parameters and the number of dimensions, and we made the dataset for this experiment.
TABLE 1  show details.

\begin{table}[h]
    \caption{data}
    \centering
    \begin{tabular}{|c|c|} \hline
      the number of data & 10,000 \\ \hline
      the number of component mixture & 4 \\ \hline
      mixing coefficient & 0.1,0.2,0.3,0.4 \\ \hline
      means, convariances& random \\ \hline
      the number of dimensions & 10 \\ \hline
    \end{tabular}
\end{table}

We measured how the FIC changed with each iteration to compare the convergence speed of online learning with it of batch learning.
The number of iterations until convergence was also included in the evaluation.
The experiment is performed 10 times, and the average is taken as the experimental result.
We experimented by changing the number of data [500, 10000] (Fig. 1) and changing the number of dimensions [2, 4, 20] (Fig. 2).
We experimented with the other parameters fixed.

\begin{figure}[]
  \begin{center} 
    \includegraphics[clip,width=8cm,height=3.3cm]{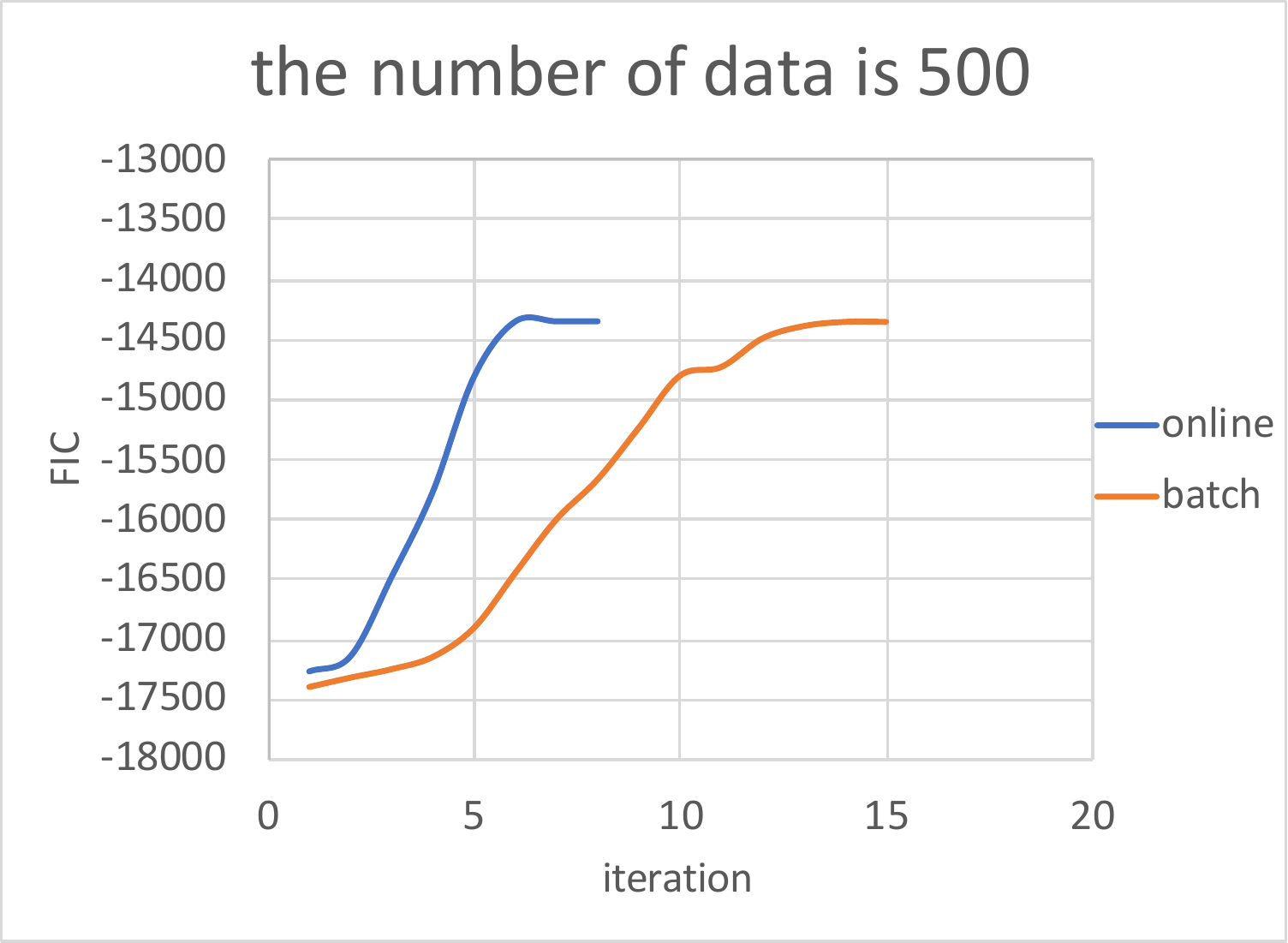}
    \includegraphics[clip,width=8cm,height=3.3cm]{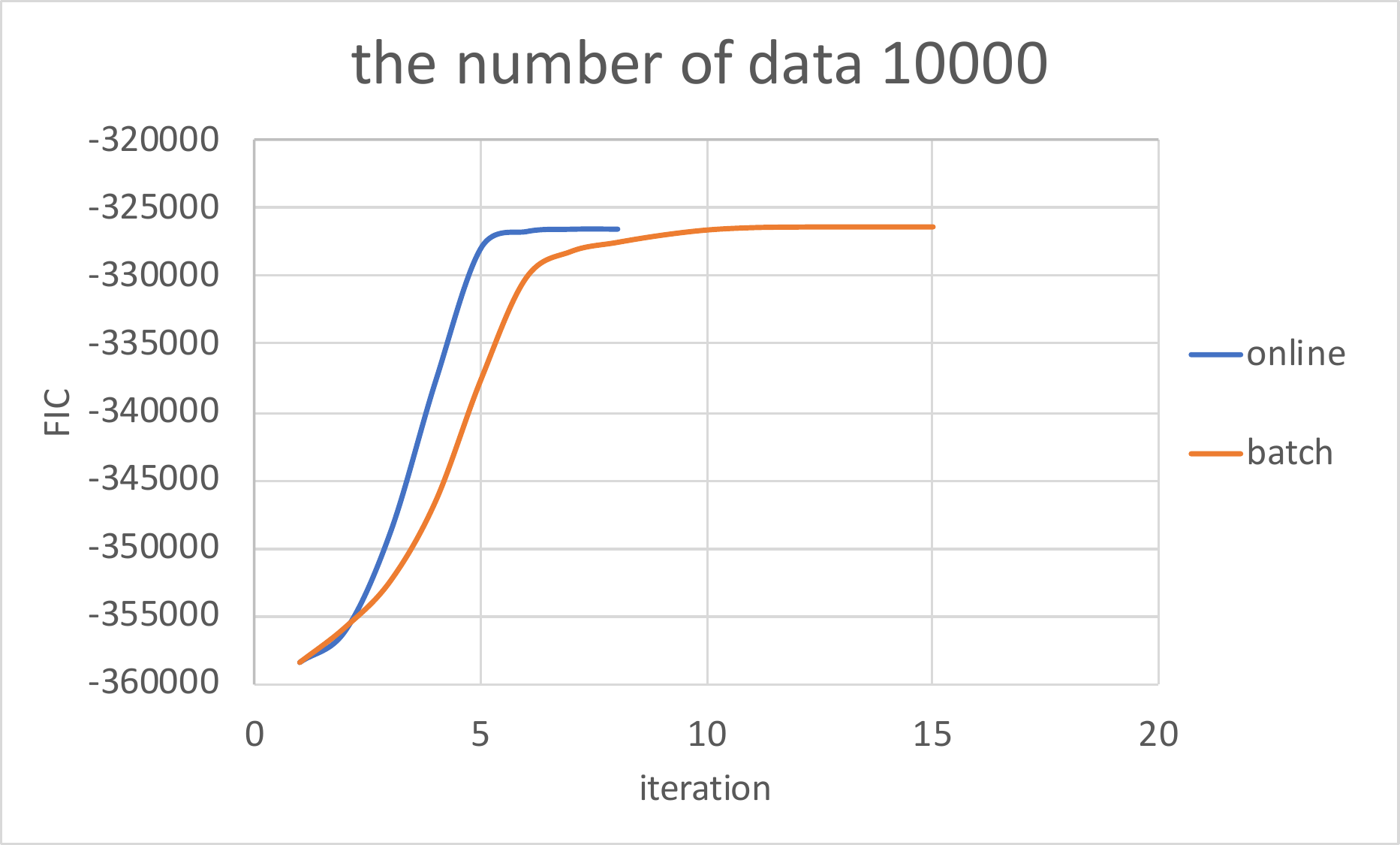}
    \caption{the change in FIC between batch and online iterations in the initial learning with the number of data changed to [500, 10000]} 
    \includegraphics[clip,width=8cm,height=3.2cm]{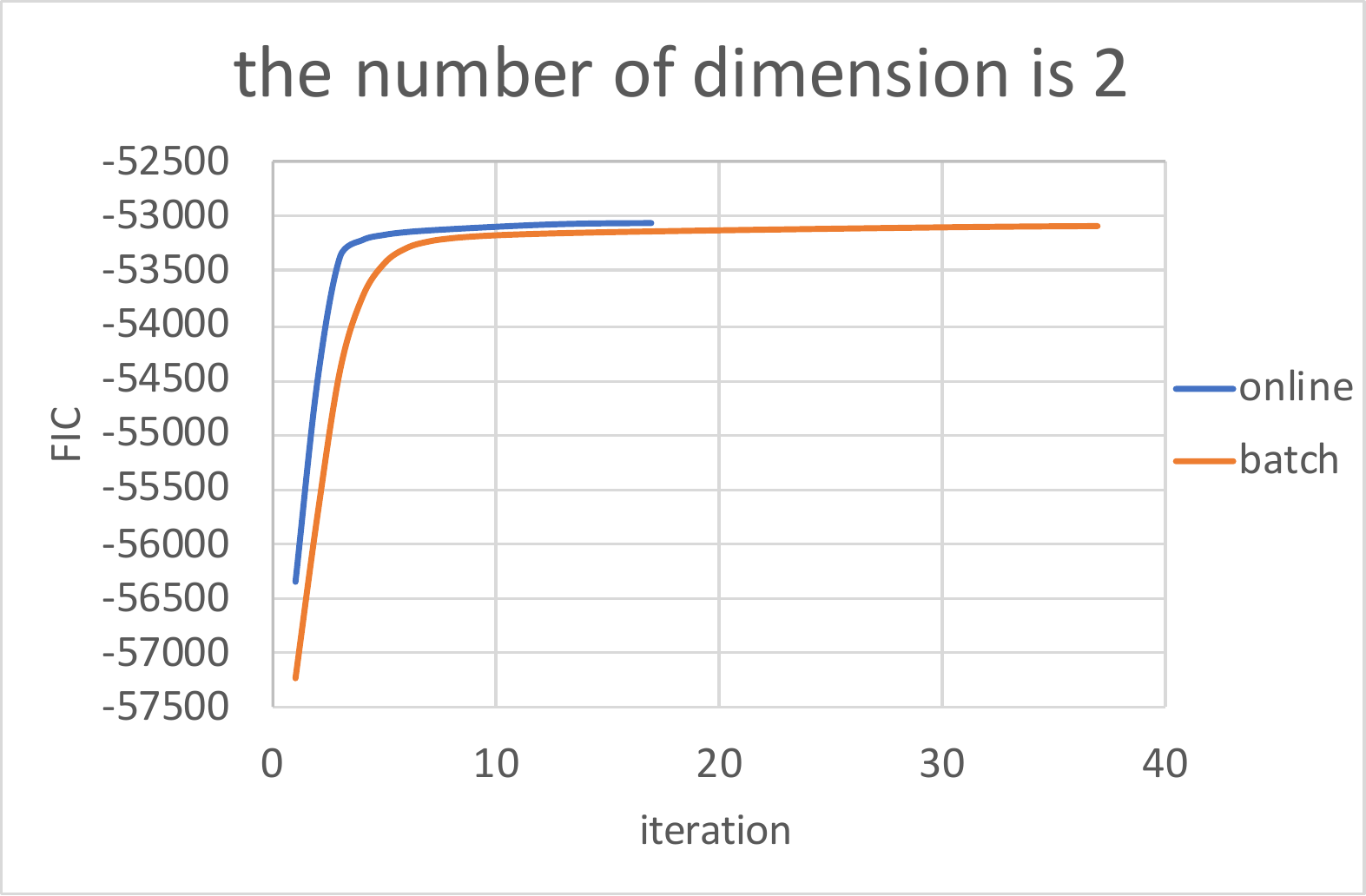}
    \includegraphics[clip,width=8cm,height=3.2cm]{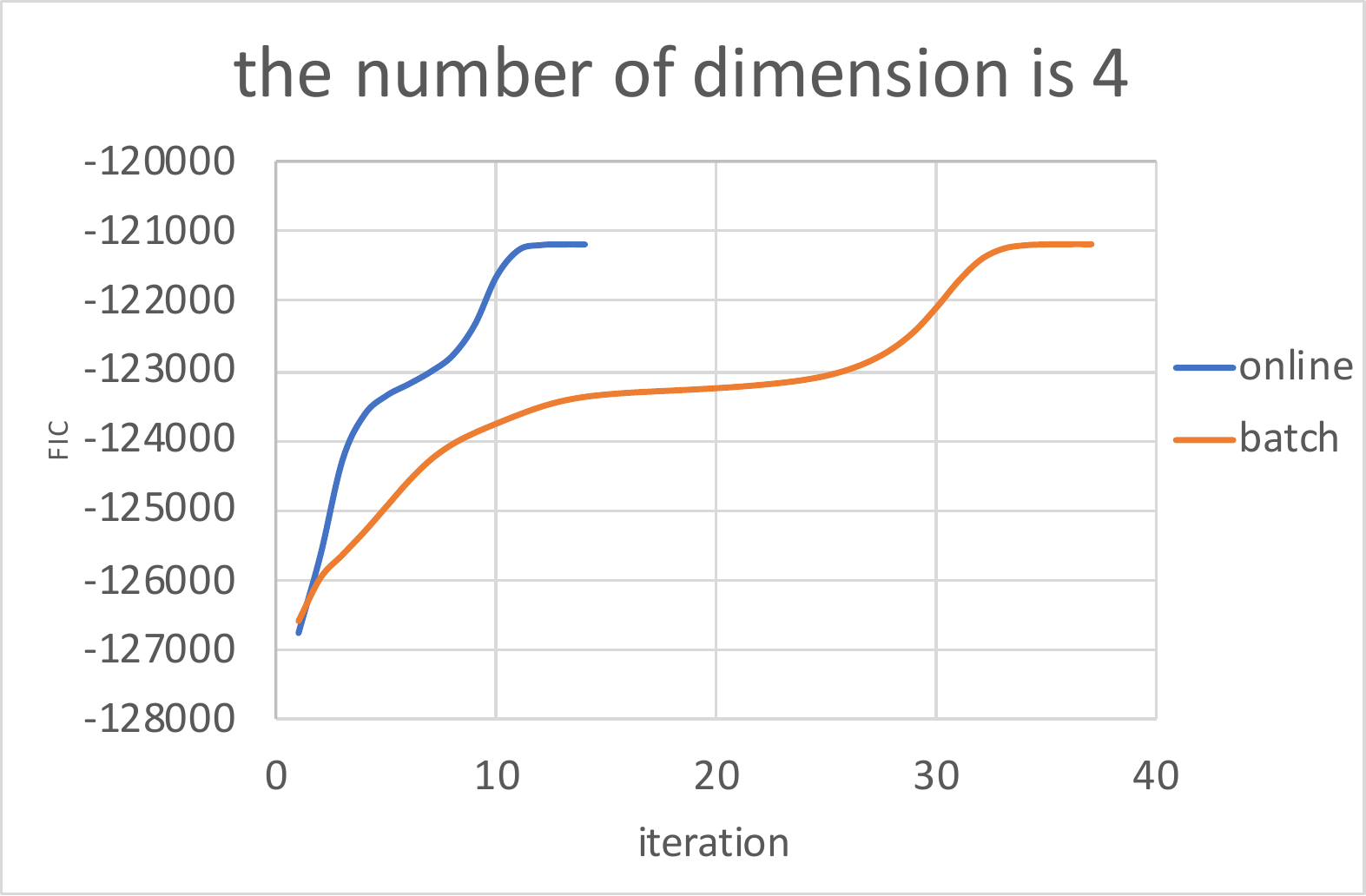}
    \includegraphics[clip,width=8cm,height=3.2cm]{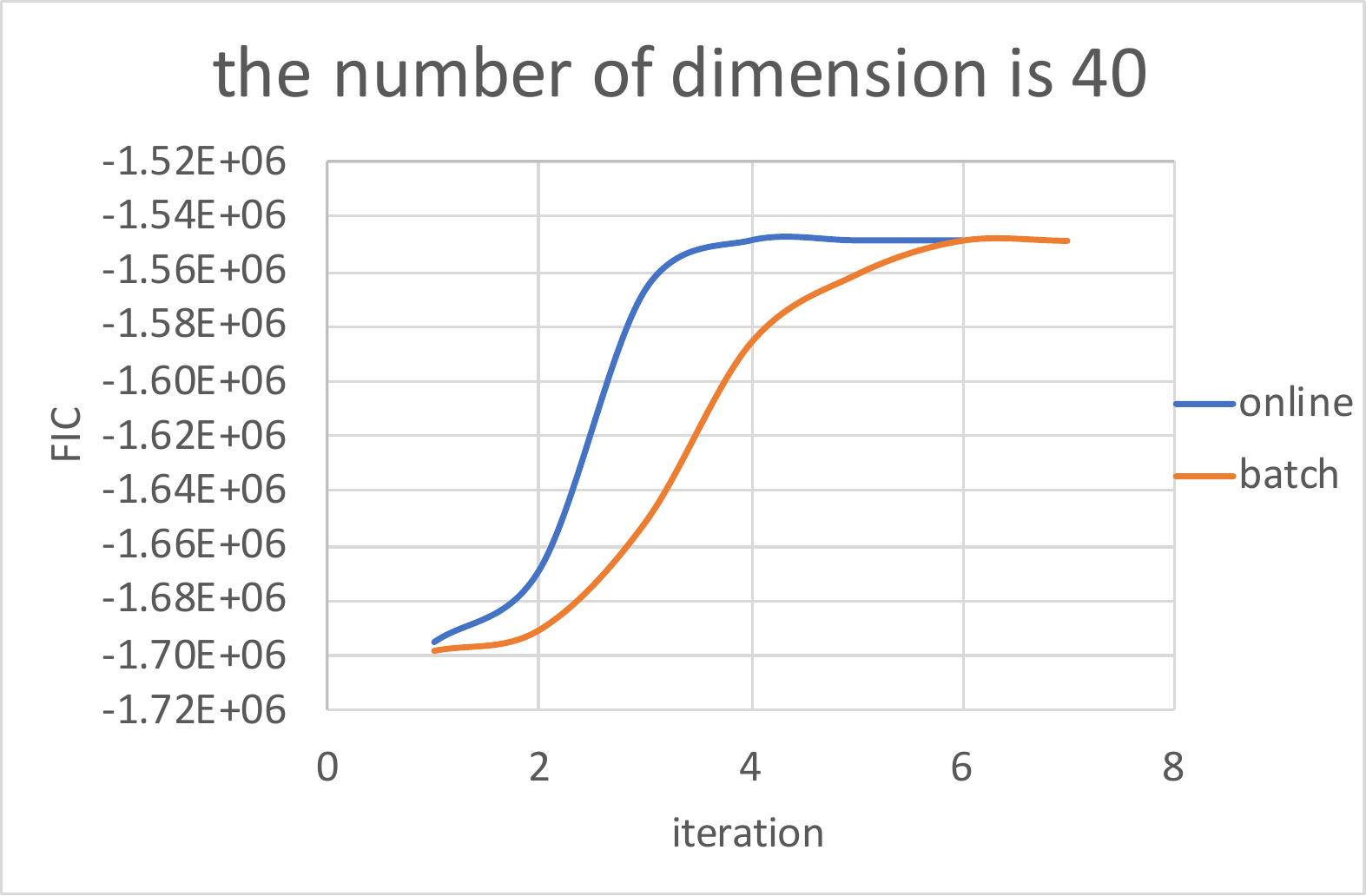}
    \caption{the change in FIC between batch and online iterations in the initial learning with the number of dimensions changed to [2, 4, 40]}
    \label{fig:wide} 
  \end{center}
\end{figure}

In terms of the small number of iterations, online learning is better than batch learning for all diagrams.
When comparing online and batch algorithms, the batch algorithm usually converges faster.
However, in the EM algorithm, the on-line algorithm converges faster than the batch algorithm.

\section{CONCLUSION}
We proposed the online heterogeneous mixture learning for the purpose of speeding up the convergence of machine learning for heterogeneous data.
It can be learned to the same accuracy of the batch one with fewer iterations than the batch one.
It is also necessary to consider using the Stepwise EM algorithm [4] in which the work area is scalable.

\end{document}